\pdfoutput=1

\documentclass[11pt]{article}

\usepackage[final]{./acl}

\usepackage{times}
\usepackage{latexsym}

\usepackage[T1]{fontenc}
\usepackage{amsfonts}

\usepackage{tabularx}

\usepackage[utf8]{inputenc}

\usepackage{microtype}

\usepackage{inconsolata}

\usepackage{graphicx}
\usepackage{float}

\title{GigaEmbeddings --- Efficient Russian Language Embedding Model}

\author{
 \textbf{Kolodin Egor\textsuperscript{1,2}},
 \textbf{Khomich Daria\textsuperscript{2,3}},
 \textbf{Savushkin Nikita\textsuperscript{2,3}},\\
 \textbf{Ianina Anastasia\textsuperscript{1,4}},
 \textbf{Fyodor Minkin\textsuperscript{1,2}}
\\
\\
 \textsuperscript{1}MIPT,
 \textsuperscript{2}SaluteDevices,
 \textsuperscript{3}MSU,
 \textsuperscript{4}Wildberries
\\
 \small{
   \textbf{Correspondence:} \href{mailto:kolodin.ei@phystech.edu}{kolodin.ei@phystech.edu}
 }
}

\begin{document}
\maketitle
\begin{abstract}
We introduce \texttt{GigaEmbeddings}, a novel framework for training high-performance Russian-focused text embeddings through hierarchical instruction tuning of the decoder-only LLM designed specifically for Russian language (\texttt{GigaChat-3B}). Our three-stage pipeline, comprising large-scale contrastive pre-training in web-scale corpora, fine-tuning with hard negatives, and multitask generalization across retrieval, classification, and clustering tasks, addresses key limitations of existing methods by unifying diverse objectives and leveraging synthetic data generation. Architectural innovations include bidirectional attention for contextual modeling, latent attention pooling for robust sequence aggregation, and strategic pruning of 25\% of transformer layers to enhance efficiency without compromising performance. Evaluated on the ruMTEB benchmark spanning 23 multilingual tasks, \texttt{GigaEmbeddings} achieves state-of-the-art results (69.1 avg. score), outperforming strong baselines with a larger number of parameters. 
\end{abstract}

\section{Introduction}

Text embeddings, vector representations that encode semantic information from natural language, serve as foundational components across diverse natural language processing (NLP) applications. These include information retrieval (IR), question answering, semantic similarity evaluation, bi-text mining, and recommendation systems. In IR pipelines, embeddings enable efficient first-stage retrieval through approximate nearest neighbor search, narrowing vast corpora to manageable candidate sets. Their role extends to retrieval-augmented generation (RAG) \cite{NEURIPS2020_RAG}, where they dynamically ground large language models (LLMs) in external knowledge, and to source attribution frameworks \cite{gao-etal-2023-enabling}, enhancing the transparency of LLM outputs.

The textual embeddings for languages other than English should be treated with even more attention. It requires careful consideration of language support and data which ideally should be large-scale and domain-specific. Thus, researching embedding models for less popular or even low-resource languages helps to improve the quality of various tasks, such as article recommendation, assessing semantic similarity, information retrieval, intent recognition and many more for different communities where English is not the main spoken language. Our study focuses on embedding models in the Russian language.

Early approaches to text embeddings, such as weighted averages of static word embeddings \cite{pennington-etal-2014-glove}, provided rudimentary semantic signals but lacked contextual awareness. The emergence of pre-trained language models \cite{devlin-etal-2019-bert} catalyzed advances like Sentence-BERT \cite{devlin-etal-2019-bert} and SimCSE \cite{gao-etal-2021-simcse}, which fine-tune BERT on natural language inference (NLI) tasks to produce context-sensitive embeddings. State-of-the-art methods, including E5 \cite{wang2024multilinguale5textembeddings} and BGE \cite{chen-etal-2024-m3}, further scale performance through multi-stage pipelines: pre-training on weakly supervised web-scale pairs followed by task-specific fine-tuning.

However, these paradigms face critical limitations. First, their multistage workflows demand labour-intensive curation of massive relevance pairs, often restricted to narrow task domains or high-resource languages - a particular challenge for Russian, which remains underserved because of its linguistic complexity and lack of dedicated embedding models. Second, reliance on BERT-style encoders ignores breakthroughs in modern LLM architectures, such as extended context windows \cite{wang2024textembeddingsweaklysupervisedcontrastive} and parameter-efficient adaptation techniques. Third, static training data fails to leverage synthetic data generation capabilities of instruction-tuned LLMs, constraining cross-lingual generalization, a critical gap for Russian-centric applications, where low-resource constraints and domain-specific nuances demand flexible, language-aware training paradigms.

In this work, we address these gaps by introducing \texttt{GigaEmbeddings}, a three-stage instruction-tuning framework built on the \texttt{GigaChat-3B} decoder-only LLM backbone — a member of the state-of-the-art \texttt{GigaChat} family \footnote{\url{https://huggingface.co/ai-sage}}, one of the most advanced Russian-language large language models developed by Salute Devices and renowned for its robust performance on Russian NLP benchmarks. Our approach (1) synthesizes diverse, multilingual training pairs via LLM-generated queries, eliminating dependency on manually curated datasets; (2) unifies retrieval, classification, and clustering objectives through dynamic task-aware instruction tuning; and (3) integrates architectural innovations such as latent attention pooling and layer pruning to enhance efficiency. By combining synthetic data scalability with the multilingual capabilities of the \texttt{GigaChat} foundation, which excels in both Russian and cross-lingual tasks, \texttt{GigaEmbeddings} achieves state-of-the-art performance across 23 multilingual tasks on the ruMTEB benchmark. This work demonstrates the viability of decoder-only LLMs as universal encoders and establishes a new paradigm for embedding model training in low-resource and multitask settings.

Our work introduces a three-stage instruction-tuning methodology for large language models (LLMs), designed to optimize performance and efficiency through pretraining, fine-tuning, and multitask learning. The pretraining phase employs contrastive learning with large batches (16,384 samples) and a combination of in-batch and cross-batch negatives, enabling robust semantic representation learning. During fine-tuning, we leverage high-quality labeled datasets and introduce hard negatives (7 per query) to sharpen discriminative capabilities. The final stage of multitask learning integrates classification and clustering tasks, enhancing the applicability of the model. To enhance computational efficiency without sacrificing performance, we prune 25\% of the original LLM’s layers, guided by insights from \citet{gromov2025unreasonableineffectivenessdeeperlayers}, also achieving a reduction in inference latency. We open source our model at HuggingFace. \footnote{\url{https://huggingface.co/ai-sage/Giga-Embeddings-instruct}}

\begin{figure*}[h!]
    \centering
    \includegraphics[width=0.8\linewidth]{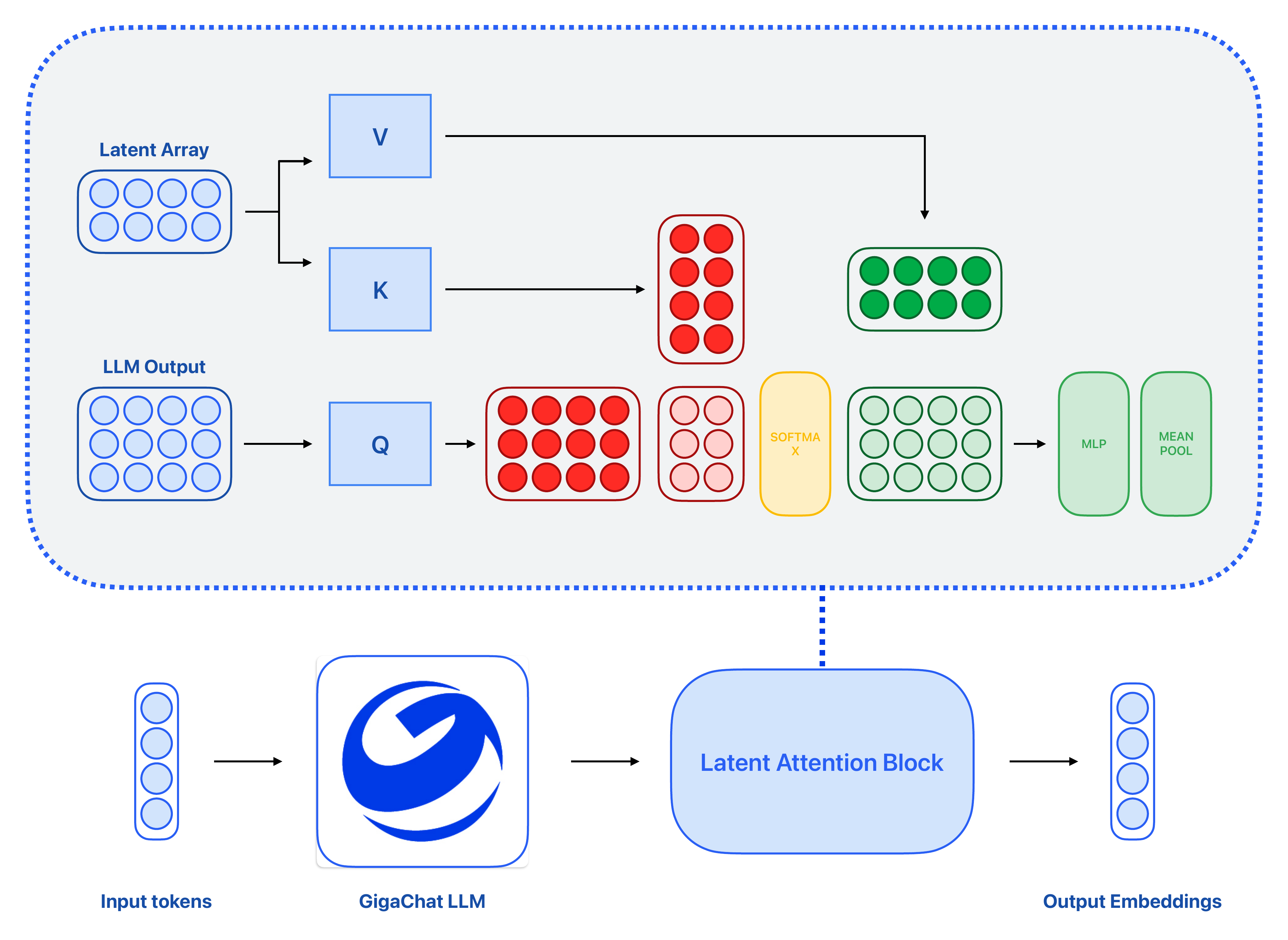} \hfill
    \caption{The design comprises a \texttt{GigaChat} LLM followed by a latent attention layer. This layer operates through a cross-attention mechanism where the decoder’s output acts as the query (Q), a trainable latent array provides the key (K) and value (V) inputs. The attention output is then processed by a multilayer perceptron (MLP).}
    \label{fig:scheme}
\end{figure*}

\section{Related work}

Recent advances in embedding models have focused on improving generalization, efficiency, and multilingual capabilities. 

\subsection{Leveraging Pretrained Language Models}

Pretrained LLMs are increasingly being adopted as the backbone for embedding models, reflecting a major shift in the field. For instance, \citet{wang-etal-2024-improving-text} adapts the Mistral decoder-only LLM \cite{jiang2023mistral7b} through contrastive fine-tuning, achieving top performance on the MTEB benchmark \cite{muennighoff-etal-2023-mteb}. Similarly, LLM2Vec \cite{behnamghader2024llmvec} transforms decoder-only models such as LLaMA \cite{touvron2023llama2openfoundation} and Mistral \cite{jiang2023mistral7b} into universal text encoders via parameter-efficient methods \cite{xu2023parameterefficientfinetuningmethodspretrained}, demonstrating that causal LLMs can rival traditional encoder-only architectures when properly adapted. These approaches align with Gecko \cite{lee2024geckoversatiletextembeddings}, which distills ranking capabilities from GPT-4 into smaller models, proving that LLM-generated relevance signals can replace human-labeled data. 

\subsection{Efficient Architectures and Long-Context Modeling}

Modern embedding models optimize efficiency and long-context handling with architectural redesign. \citet{Nussbaum2024NomicET} modifies BERT with rotary positional embeddings, Flash Attention, and Dynamic NTK interpolation to scale to 8k tokens. \citet{sturua2024jinaembeddingsv3multilingualembeddingstask} employs task-specific LoRA adapters \cite{xu2023parameterefficientfinetuningmethodspretrained} and Matryoshka Representation Learning \cite{Matryoshka}, enabling dimension reduction (1024 to 32) without performance loss. \citet{merrick2024arctic} focuses on dataset stratified batching and hard negative mining to improve training efficiency.

\subsection{Synthetic Data for Training Embedding Models}

The use of synthetic data to train embedding models has emerged as a critical strategy to address data scarcity and improve generalization. Recent work shows that LLMs can generate high-quality training pairs in diverse tasks and languages. For example, \citet{wang-etal-2024-improving-text} introduces a taxonomy-driven approach to generate task-specific synthetic data. By categorizing tasks into asymmetric (e.g. short-long query-document matches) and symmetric (e.g. semantic textual similarity) groups, the authors design structured prompt templates with randomized placeholders to maximize diversity. Their two-step prompting method: first brainstorming task definitions, then generating examples ensures coherence while scaling across 93 languages.

Further innovations refine synthetic data quality through iterative distillation and filtering. \citet{lee2024geckoversatiletextembeddings} employs a two-step distillation process: Initial synthetic query-document pairs are generated by an LLM, followed by re-labeling hard negatives and positives using the same LLM to improve relevance signals. This approach enables Gecko to outperform larger models on the MTEB benchmark \cite{muennighoff-etal-2023-mteb}. Similarly, \citet{merrick2024arctic} generates synthetic queries while grounding them with real negative documents, observing that LLMs struggle to produce high-quality negatives independently. Their hybrid strategy, combining LLM-generated queries with mined negatives, yields significant performance gains, as validated by HotpotQA \cite{yang-etal-2018-hotpotqa} evaluations.

For specialized tasks such as long document and multilingual retrieval, \citet{chen-etal-2024-m3} augments training data by sampling lengthy articles from Wikipedia and mC4 \cite{raffel2023exploringlimitstransferlearning}, then synthesizing questions through GPT-3.5. This task-targeted generation mitigates data shortages in long-context scenarios while improving cross-lingual alignment.

\section{Method}

This section details our methodology, beginning with the core training objective, followed by the three-stage instruction-tuning pipeline, and concluding with architectural optimizations.

\subsection{Training Objective}
We employ the classic InfoNCE loss \cite{oord2019representationlearningcontrastivepredictive} with a fixed temperature \( \tau = 0.02 \):  
\[
\min \mathcal{L} = -\log \frac{\phi(q, d^+)}{\phi(q, d^+) + \sum\limits_{n_i \in \mathbb{N}} \phi(q, n_i)},
\]  

where $\mathbb{N}$ denotes the set of all in-batch, cross-batch, and hard negatives, and $\phi(q, d^+)$ is a function that computes the matching score between query $q$ and positive document $d^+$. In this paper, we use the temperature-scaled cosine similarity function as follows:

\[
\phi(q, d^+) = \exp({\frac{1}{\tau}} \cos (\textbf{h}_q, \textbf{h}_{d^+}))
\]   

And $\textbf{h}_q, \textbf{h}_{d^+}$ are vector embeddings obtained from the model. 

\subsection{Three-Stage Instruction-Tuning}
The choice of the \texttt{GigaChat-3B} pretrain checkpoint as our backbone model reflects a deliberate balance between computational efficiency and model capability. With limited access to large-scale GPU clusters (e.g., 8×A100 80GB nodes), we prioritized a parameter scale that enables efficient fine-tuning and inference while retaining sufficient representational power for complex Russian-language tasks — a critical consideration given the scarcity of dedicated Russian embedding models. Building upon this foundation, we design a hierarchical three-stage instruction-tuning pipeline to progressively refine embeddings for diverse downstream tasks, ensuring adaptability to both low-resource constraints and multilingual applications.

\subsubsection{Pretraining with Large-Batch Contrastive Learning}  

The first stage leverages web-scale corpora in title-passage format sourced from Wikipedia, Reddit, StackExchange, and S2ORC \cite{lo-etal-2020-s2orc}, alongside mined raw text passages. To enhance query diversity, we employ an instruction-tuned LLM to synthetically generate contextually relevant queries for each passage. Training utilizes the InfoNCE loss with in-batch negatives (sampled from the same batch) and cross-batch negatives (cached from recent batches), enabling efficient contrastive learning across 16,384 samples per batch. This large batch size maximizes negative example diversity while adapting the decoder-only \texttt{GigaChat} LLM architecture to bidirectional attention for embedding tasks.

\subsubsection{Fine-Tuning with Hard Negatives}

The second stage focuses on high-quality labeled datasets, including MS-MARCO \cite{bajaj2018msmarcohumangenerated}, Natural Questions (NQ) \cite{kwiatkowski-etal-2019-natural}, SQuAD \cite{rajpurkar-etal-2016-squad}, MIRACL \cite{zhang-etal-2023-miracl}, and Mr TyDi \cite{zhang-etal-2021-mr}, to refine retrieval-specific capabilities. We extend the InfoNCE loss by incorporating 7 curated hard negatives per query, mined via semantic similarity thresholds, alongside in-batch negatives. The batch size is reduced to 512 samples to prioritize precision over scale, ensuring robust discriminative training for challenging retrieval scenarios. 

\subsubsection{Multitask Generalization}

The final stage introduces classification and clustering tasks into the training mixture of retrieval, classification and clustering tasks to broaden the applicability of the model. To prevent false negatives in non-retrieval tasks, we remove in-batch negatives with task-specific instruction tuning, employing a unified InfoNCE loss across all objectives. A batch size of 512 samples proves sufficient to balance computational efficiency with performance gains on classification and clustering benchmarks. 

Our methodology employs an example-driven sampling strategy: positive instances are selected from the same class/cluster as the query example, while negative instances are drawn from different classes/clusters. This approach ensures intraclass cohesion and interclass distinction during training. \cite{lee2025nvembed}

This hierarchical approach, from large-scale pretraining to task-specialized fine-tuning and multitask generalization, ensures that the model first learns broad semantic patterns, then hones discriminative precision and finally achieves cross-task robustness. 

\subsection{Architectural Innovations}

\textbf{Bidirectional Attention and Layer Pruning.}
We remove causal attention masks during training to enable bidirectional context modeling. Furthermore, inspired by \citet{gromov2025unreasonableineffectivenessdeeperlayers}, we prune 25\% of deeper transformer layers, reducing the computation with negligible quality loss.  

\textbf{Latent Attention Pooling.}
Inspired by \citet{lee2025nvembed}, we use a latent attention pooling head to process hidden activations to obtain the final embedding vector, since it works the best in our experiments. The model diagram is shown in Figure~\ref{fig:scheme}.

\section{Experiments}

\begin{table*}[h!]
  \scalebox{0.96}{
      \begin{tabular}{lccccccc}
        \hline
        \textbf{Model} & Class. & Cluster. & MultiClass. & PairClass. & Rerank & Retrieval & STS \\
        \hline
        \texttt{e5-large-instruct} & 66.28 & 63.13 & 41.15 & \textbf{63.89} & 64.35 & 68.23 & \textbf{76.48} \\
        \texttt{e5-mistral-7b-instruct} & 69.07 & 64.24 & 42.93 & 60.81 & 69.96 & 74.19 & 73.71 \\
        \texttt{SFR-Embedding-Mistral} & 69.81 & 64.92 & 42.95 & 60.65 & 70.46 & - & 74.31 \\
        \texttt{GritLM-7B} & 69.92 & 64.3 & 41.96 & 58.93 & 69.99 & \textbf{75.79} & 74.63 \\
        \texttt{BGE-M3} & 60.44 & 52.38 & 34.86 & 60.6 & 69.71 & 74.79 & 73.68 \\
        \texttt{GigaEmbeddings} & \textbf{72.7} & \textbf{65.36} & \textbf{51.75} & 57.85 & \textbf{73.42} & 74.28 & 72.11 \\
        \hline
    
        \hline
      \end{tabular}
  }
  \caption{\label{rumteb_results} Comparison with baselines on ruMTEB benchmark. }
\end{table*}

\subsection{Baselines}
We compare our model with several strong baselines, according to ruMTEB benchmark.
\begin{itemize}
    \item \texttt{multilingual-E5-large-instruct}~\cite{wang2022text};
    \item \texttt{E5-Mistral-instruct}~\cite{wang-etal-2024-improving-text};
    \item \texttt{SFR-Embedding-Mistral}~\cite{SFRAIResearch2024};
    \item \texttt{GritLM-7B}~\cite{muennighoff2025generativerepresentationalinstructiontuning};
    \item \texttt{BGE-M3}~\cite{chen-etal-2024-m3}. 

\end{itemize}

\subsection{Main results}
We assess the effectiveness of our model utilizing the ruMTEB benchmark \citep{snegirev2025russianfocusedembeddersexplorationrumteb} over 23 distinct tasks. Table~\ref{rumteb_results} provides an overview of the average ruMTEB scores for seven sub-categories, comparing them with leading models from the ruMTEB leaderboard. Our model, referred to as \texttt{GigaEmbeddings}, achieves a score of $69.1$, securing the top position on the ruMTEB as of December 2024 (comprehensive benchmark results can be found in Table~\ref{rumteb_results}). In the subsequent sections, we will detail pruning and ablation studies focusing on decisions regarding the model architecture, training methodology, and data selection strategy.

We evaluated our \texttt{GigaEmbeddings} against the latest front-edge embedding models using quantitative leaderboard evaluations. The \texttt{e5-mistral-7b-instruct} \cite{wang-etal-2024-improving-text} is trained with proprietary synthetic data in a single stage. Conversely, we acknowledge that retrieval tasks pose more challenges than other embedding tasks; therefore, we focus our training strategy on pre-training and fine-tuning our model for retrieval initially. Subsequently, we integrate the other subtasks into instruction-tuning within a multitask learning framework, resulting in significantly enhanced MTEB performance.

\texttt{SFR-Embedding-Mistral} \cite{SFRAIResearch2024} demonstrates competitive scores on the ruMTEB benchmark by continuing to finetune the \texttt{e5-mistral-7b-instruct} model \cite{wang-etal-2024-improving-text}. However, it remains largely constrained by the architectural limitations of its parent model, such as the causal attention mask and the last token pooling method.

\texttt{GritLM-7B}~\cite{muennighoff2025generativerepresentationalinstructiontuning}~---~a generative representational instruction tuned language model. It unifies embeddings and text generation into a single model achieving state-of-the-art performance on both types of tasks.

\texttt{BGE-M3}~\cite{chen-etal-2024-m3}~---~an embedding model, that provides a uniform support for the semantic retrieval of more than 100 languages. It can simultaneously accomplish dense retrieval, multi-vector retrieval, and sparse retrieval. 

\subsection{Pruning}
We follow the findings \citet{gromov2025unreasonableineffectivenessdeeperlayers} and prune 25\% of the last blocks of our LLM model. We remove 9 out of 36 transformer blocks -- self-attention and feed forward modules. And use this 2.5B LLM as the backbone of our embedding model. Table \ref{rumteb_results_pruned} presents the ablation.

\begin{table}[h!]
  \begin{tabular}{lc}
    \hline
    \textbf{Model} & ruMTEB \\
    \hline
    \texttt{GigaEmbeddings} \textit{full} & 69.3 \\
    \texttt{GigaEmbeddings} \textit{pruned} & 69.1 \\
    \hline

    \hline
  \end{tabular}
  \caption{\label{rumteb_results_pruned} Comparison of original and pruned versions of the models. }
\end{table}

\subsection{Ablation studies}
Since there are different types of tasks: symmetric (classification, clustering, STS) and asymmetric (retrieval, re-ranking), it was shown that the prompting model can benefit quality. We compare two ways of prompting model: prefix (\citealp{Nussbaum2024NomicET}, \citealp{wang2024multilinguale5textembeddings}) and instruct (\citealp{behnamghader2024llmvec}, \citealp{lee2025nvembed}). Table \ref{rumteb_instruct_ablation}.

\begin{table}[h!]
  \begin{tabular}{lc}
    \hline
    \textbf{Prompting strategy} & ruMTEB \\
    \hline
    \texttt{GigaEmbeddings} \textit{prefix} & 68.5 \\
    \texttt{GigaEmbeddings} \textit{instruct} & 69.3 \\
    \hline

    \hline
  \end{tabular}
  \caption{\label{rumteb_instruct_ablation} Comparison between prefix and instruct ways of model prompting. }
\end{table}

The effectiveness of current text embedding models is substantially attributed to the use of weakly supervised contrastive pre-training \cite{wang2024multilinguale5textembeddings}. Similarly to \citet{wang-etal-2024-improving-text}, we decided to check the necessity of contrastive pre-training. We compared two ways of training models: three-stage training (with pre-training at the beginning) and two-stage training (fine-tune and multitask). Table \ref{rumteb_pretrain_ablation}.

\begin{table}[h!]
  \begin{tabular}{lc}
    \hline
    \textbf{Training strategy} & ruMTEB \\
    \hline
    \texttt{GigaEmbeddings} \textit{w/ pre-training} & 69.3 \\
    \texttt{GigaEmbeddings} \textit{w/o pre-training} & 68.7 \\
    \hline

    \hline
  \end{tabular}
  \caption{\label{rumteb_pretrain_ablation} Effectiveness of pre-training stage. }
\end{table}

It was observed that including contrastive pre-training on weakly supervised data can further boost quality on retrieval tasks. We explain this by the fact that the model needs to reconfigure itself for the new task and also adapt to the new mechanism of encoder attention.

\section{Conclusion}

We presented \texttt{GigaEmbeddings}, a novel three-stage instruction-tuning framework to train high-performance text embeddings using the \texttt{GigaChat-3B} decoder-only LLM as its backbone. Our hierarchical pipeline, which spans large-scale contrastive pretraining, hard-negative fine-tuning, and multitask generalization, addresses critical limitations of existing methods by unifying retrieval, classification, and clustering objectives while leveraging synthetic data generation. Architectural innovations, including bidirectional attention and layer pruning (25\% reduction in parameters).

Evaluation on the ruMTEB benchmark demonstrates state-of-the-art performance, with \texttt{GigaEmbeddings} achieving an average score of 69.1 on 23 tasks, outperforming strong baselines like \texttt{e5-mistral-7b-instruct} and \texttt{SFR-Embedding-Mistral}. Ablation studies validate the necessity of contrastive pretraining (3.4\% gain in retrieval tasks with lower parameters) and the superiority of instruction-based prompting over prefix-based approaches. In particular, strategic pruning based on insights from \citet{gromov2025unreasonableineffectivenessdeeperlayers} retains the full model performance while significantly enhancing efficiency.

By open-sourcing our model we provide a reproducible foundation for future research. Our work establishes decoder-only LLMs as versatile encoders and highlights the potential of synthetic data-driven training paradigms. Future directions include extending this framework to low-resource languages and integrating dynamic context window scaling for long-document applications.

\section*{Limitations}
While \texttt{GigaEmbeddings} demonstrate strong performance on Russian and English tasks, several limitations merit consideration. First, the model’s multilingual capabilities are currently restricted to these two languages, as we did not evaluate its effectiveness on other languages, particularly low-resource ones. This narrow focus limits its applicability in truly global multilingual settings. Second, the computational demands of the 3B-parameter architecture and high-dimensional embeddings (e.g., 2048 dimensions) necessitate GPU acceleration, rendering deployment on CPU-only systems impractical for real-time applications. Third, while our model achieves competitive results on retrieval tasks, indicating room for improvement in dense retrieval efficiency.

These limitations highlight key directions for future work: expanding language coverage through targeted multilingual training, optimizing model size via quantization or distillation, and refining retrieval-specific architectures to bridge the performance gap.

\section*{Acknowledgments}

We would like to thank Alex Abramov, Artem Snegirev and other team members who participated in this research. 

\bibliography{custom}

\appendix

\section{Implementation Details}
\label{sec:appendix}

We list the hyperparameters in Table \ref{hyperparams}.

\begin{table}[h!]
  \scalebox{0.9}{
  \begin{tabular}{lccc}
    \hline
     & pre-training & fine-tuning & multitask \\
    \hline
    learning rate & 1e-5 & 1e-5 & 1e-5 \\
    warmup steps & 1000 & 500 & 500 \\
    batch size & 16K & 512 & 512 \\
    max steps & 6000 & n.a. & n.a. \\
    max length & 512 & 512 & 512 \\
    epochs & n.a. & 3 & 3 \\
    $\tau$ & 0.02 & 0.02 & 0.02 \\
    weight decay & 0.01 & 0.01 & 0.01 \\
    hard negatives & 0 & 7 & 7 \\
    \hline

    \hline
  \end{tabular}
  }
  \caption{\label{hyperparams} Hyperparameters for contrastive pre-training, fine-tuning and multitask stages. }
\end{table}

\end{document}